\documentclass[cameraready]{Interspeech}

\usepackage[capitalise]{cleveref}
\usepackage{xspace}
\newcommand{\ours}{\textbf{\texttt{Referee}}\xspace}
\newcommand{\idmodule}{IDB\xspace}
\newcommand{\tgt}{\text{TGT}}
\newcommand{\refer}{\text{REF}}
\newcommand{\ID}{\text{ID}}
\newcommand{\RF}{\text{RF}}
\newcommand{\eg}{\textit{e.g.}}
\newcommand{\ie}{\textit{i.e.}}
\newcommand{\tparagraph}[1]{\noindent\textbf{#1}}

\title{Referee: Reference-aware Audiovisual Deepfake Detection}

\author[affiliation={1}, equalcontribution]{Hyemin}{Boo}
\author[affiliation={1}, equalcontribution]{Eunsang}{Lee}
\author[affiliation={1}, correspondingauthor]{Jiyoung}{Lee}

\address{
    $^1$ Division of Artificial Intelligence \& Software, Ewha Womans University
}

\email{hyeminb@ewha.ac.kr, leeeunsang@ewha.ac.kr, lee.jiyoung@ewha.ac.kr}

\keywords{deepfake detection, speaker verification}

\usepackage{tabularx}
\usepackage{amsmath}
\usepackage{makecell}
\usepackage{array} 

\usepackage{comment}
\usepackage{color}
\usepackage{colortbl}
\usepackage{tabu}
\usepackage{multirow}
\usepackage{pifont}
\usepackage{amssymb}

\definecolor{myorange}{RGB}{255, 208, 0} 
\definecolor{mygreen}{RGB}{46, 75, 191} 
\definecolor{light_gray}{gray}{0.7}

\newcommand{\cmark}{\ding{51}}%
\newcommand{\xmark}{\ding{55}}%

\begin{document}
\maketitle


\begin{abstract}
Deepfakes generated by advanced generative models have rapidly posed serious threats, yet existing audiovisual deepfake detection approaches struggle to generalize to unseen manipulation methods.
To address this, we propose a novel reference-aware audiovisual deepfake detection method, called \ours, to capture fine-grained identity discrepancies.
Unlike existing methods that overfit to transient spatiotemporal artifacts, \ours employs identity bottleneck and matching modules to model the relational consistency of speaker-specific cues captured by a single one-shot example as a biometric anchor.
Extensive experiments on FakeAVCeleb, FaceForensics++, and KoDF demonstrate that \ours achieves state-of-the-art results on cross-dataset and cross-language evaluation protocols, including a 99.4\% AUC on KoDF.
These results highlight that explicitly correlating reference-based biometric priors is a key frontier for achieving generalized and reliable audiovisual forensics.
\end{abstract}

\section{Introduction}\label{sec:intro}
Rapid advances in generative AI allow models~\cite{kowalski2018faceswap, jia2018transfer, flux2024} to synthesize high-fidelity multimodal content, including video and audio.
Despite its creative potential, this capability poses significant risks of misuse and societal harm, exemplified by \textit{deepfake}~\cite{tolosana2020deepfakes} that convincingly imitate humans.
Deepfake detection aims to distinguish genuine (\ie, real) or spoof (\ie, fake) content.
Prior unimodal approaches have primarily focused on image-based artifacts, including shape distortion~\cite{guo2022eyes, li2018exposing}, boundary artifacts~\cite{li2020face, chen2023watching,fu2025faces}, or frequency anomalies~\cite{tan2024frequency, li2021frequency}.
Alternatively, audio-based methods~\cite{qais2022deepfake,chakravarty2024lightweight} have targeted discriminative spectral patterns to identify synthetic speech. 
Although these methods have shown promising performance, the synthetic content is generated with higher fidelity, making unimodal detection strategies less reliable.

In contrast, audiovisual deepfake detection~\cite{agarwal2020detecting, liu2024lips, zhou2021joint, oorloff2024avff} has exploited the natural correspondence between facial movements and speech.
These methods have investigated dynamic motion patterns~\cite{liang2024speechforensics, yang2023avoid}, such as lip synchronization and temporal transitions, to capture manipulation cues.
However, because these approaches rely on fine-grained visual cues within the lip area, they remain vulnerable to partial lip occlusions caused by hands, microphones, or head poses.
Furthermore, their effectiveness diminishes significantly in low-resolution conditions where such subtle spatial details are heavily degraded.



When evaluating video authenticity, humans do not fixate on lip motion in isolation; instead, they holistically integrate diverse cues, \eg, tone of voice, facial appearance, and identity coherence.
Prior identity-based methods like ID-Reveal~\cite{idreveal} and ICT-Ref~\cite{ictref} rely on visual-only comparisons, often employing proxy tasks with heavy generative models or simple mask-based transformation. 
Such reliance can cause these models to overfit to low-level synthesis artifacts rather than capturing identity features. 
POI-Forensics~\cite{poiforensics} incorporates audiovisual data with na\"ive feature-level concatenation for contrastive learning, which precludes meaningful crossmodal exchange.
This structural limitation not only prevents the model from capturing intricate cross-modal dependencies but also necessitates excessively long reference footage (\eg, over five minutes per identity), rendering it impractical for real-world deployment.


\begin{figure*}[t]
    \centering
    \includegraphics[width=.9\linewidth]{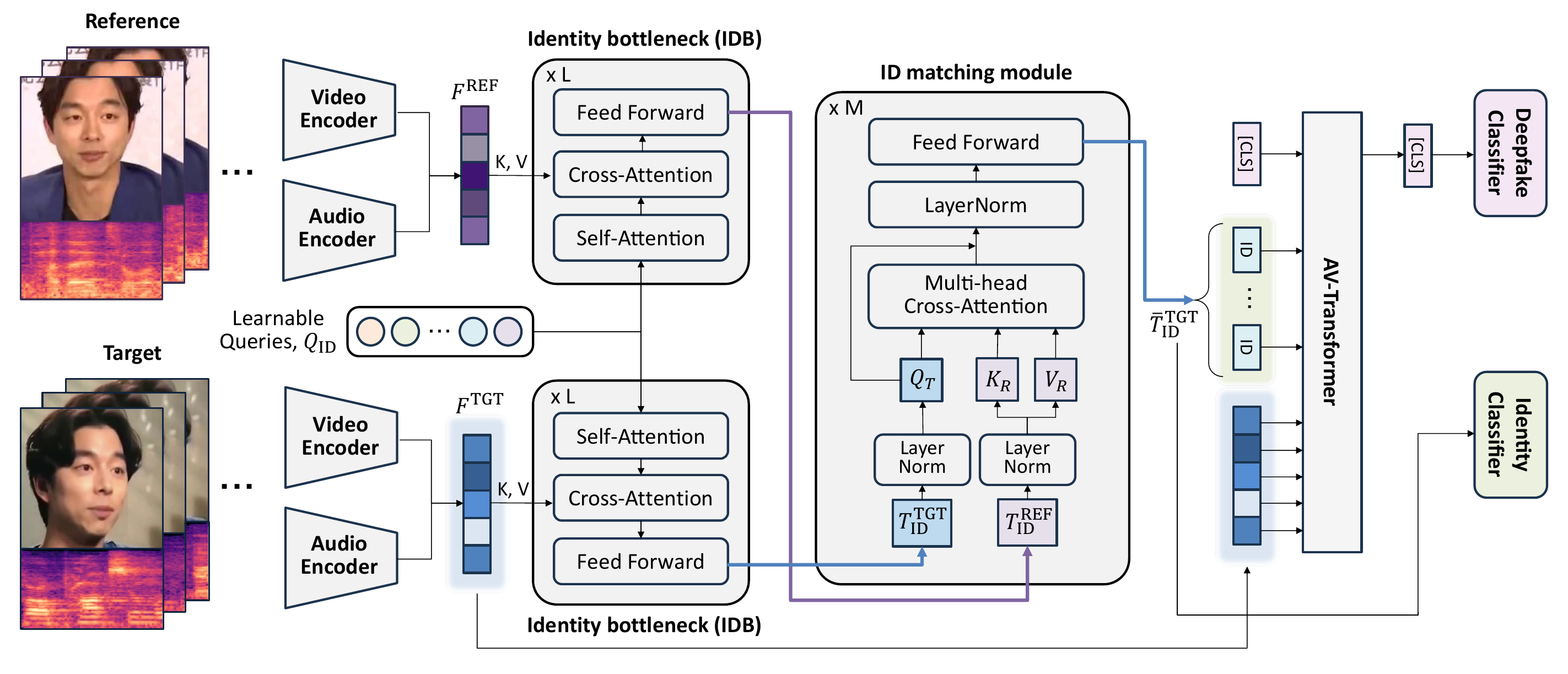}
    \vspace{-1em}
    \caption{The overall framework of \ours. 
    Identity queries are extracted by the \idmodule and conditionally refined in the ID matching module relative to the reference.
    These reference-aware queries, alongside target features, are processed by the AV-Transformer for deepfake classification, while the average-pooled reference-aware queries are used for identity classification.
    }
    \label{fig:framework}\vspace{-1em}
\end{figure*}

In this paper, we introduce \ours, a novel framework that jointly captures speaker identity inconsistency and audiovisual desynchronization for robust deepfake detection.
\ours presents an identity bottleneck module (\idmodule) with learnable identity queries to encode audiovisual sequences into a unified identity representative feature.
Identity queries obtained from both reference and target content are subsequently aligned through cross-attention to verify speaker identity.
This process is optimized by our ID matching loss to ensure that the models can capture speaker-specific cues rather than superficial artifacts.
The fused identity representation is then attached to the audiovisual feature, enabling the system to reason jointly about two critical aspects: (1) whether lip-speech synchronization is consistent over time, and (2) whether either or both modalities have been manipulated in a way that violates identity integrity.
Finally, the model integrates these signals to classify the content as real or fake.
Extensive experiments show \ours effectively surpasses current state-of-the-art works~\cite{oorloff2024avff, nie2024frade, anshul2025intra} on FakeAVCeleb~\cite{khalid2021fakeavceleb}, FaceForensics++~\cite{rossler2019faceforensics++}, and KoDF~\cite{kwon2021kodf}.
It also demonstrates that \ours is inherently more resilient to the rapid progress of generative models and better equipped to handle unseen types of deepfakes.
To sum up, our contributions are: 
\begin{itemize}
    \item We propose \ours, a reference-aware deepfake detection method that leverages audiovisual correspondence in both identity- and temporal-levels.\vspace{-.3em}
    \item \ours employs a novel identity bottleneck module with learnable queries and a cross-identity matching mechanism, enabling robust reasoning for audiovisual integrity.\vspace{-.3em}
    \item Extensive experiments on FakeAVCeleb, FaceForensics++, KoDF demonstrate that our \ours achieves state-of-the-art performance on both unseen forgery manipulations and cross-lingual evaluation. \vspace{-.7em}
\end{itemize}

\section{Method}
\vspace{-.3em}
\subsection{Motivation and overview}\vspace{-.3em}
Most prior approaches~\cite{agarwal2020detecting, liu2024lips, zhou2021joint, oorloff2024avff} for audiovisual deepfake detection have shown impressive performance while concentrating on correct synchronization between audio and video streams.
However, there are several challenges: (1) methods relying on synchronization are inherently tied to lip-sync manipulation methodologies, making it difficult to generalize to unseen forgeries.
(2) These methods also remain vulnerable under occlusion, low resolution, or degraded signals.
To solve these limitations, we propose \ours to detect forgery with speaker identity information.
We claim that cross-modal biometric cues~\cite{nagrani2018seeing, lee2023imaginary}, such as verifying the correspondence between the speaker's voice and facial appearance, provide a strong basis for deepfake detection.

Given a target and reference video including audio, \ours detects subtle identity inconsistencies within the same individual.
The proposed method mainly consists of three parts; first, we propose an identity bottleneck module, \idmodule, to capture the cross-modal identity cues into learnable ID queries.
Secondly, identity matching layers learn to check the consistency of speaker identity with the compressed ID query tokens, where the target tokens are re-aligned from the perspective of the reference identity.
Finally, the refined ID query is concatenated with the audiovisual feature, and jointly learned within the AV-Transformer to classify the video as real or fake.
The overall framework is illustrated in \cref{fig:framework}.
\vspace{-.5em}

\subsection{Audiovisual feature extraction}\vspace{-.3em}
Given an input video and its corresponding audio, our framework first extracts modality-specific representations.
To leverage robust temporal alignment, our visual and audio encoders are built on pretrained audiovisual encoders~\cite{iashin2024synchformer}.
The input streams are divided into $N_{seg}$ fixed-length segments, each with a duration of $T$ seconds.
For each segment, the encoders yield segment-level token embeddings, $F_v\in \mathbb{R}^{(N_{\text{seg}} \cdot T_v) \times D}$ and $F_a\in \mathbb{R}^{(N_{\text{seg}} \cdot T_a) \times D}$, where $T_v$ and $T_a$ denote the number of visual and audio tokens per segment, and $D$ is the feature dimension. 
To facilitate cross-modal interaction, the visual and audio sequences are concatenated to form a joint representation for both the target and reference samples.
With a learnable modality-separation token $F_{\text{mod}}$, the joint audiovisual sequences are constructed: 
\begin{equation}
F^{\tgt} = [F_{v}^{\tgt};\,F_{\text{mod}};\,F_{a}^{\tgt}], 
F^{\refer} = [F_{v}^{\refer
};\,F_{\text{mod}};\,F_{a}^{\refer}].
\end{equation} 

\vspace{-.5em}
\subsection{Identity bottleneck (\idmodule)}\vspace{-.3em}
A fundamental challenge in robust identity (ID) compression is that speaker identity is often inherently entangled with transient factors like speech content, facial expression, and head poses. 
To isolate identity-relevant signals from these confounding variables, we introduce a new ID bottleneck (\idmodule).
Our \idmodule employs a set of $N_q$ learnable queries~\cite{jaegle2021perceiver,li2023blip}, $Q_{\text{ID}}\in \mathbb{R}^{N_q\times D}$, to extract and summarize cross-modal identity cues. 
Crucially, we employ a weight-sharing scheme for the \idmodule across both target and reference branches.
The module consists of $L$ identical transformer blocks, each composed of a self-attention layer, a cross-attention layer, and a feed-forward network. 
To stabilize learning and encode relative positions, each query is initialized with a learnable positional embedding.

Based on the cross-attention mechanism, the learnable queries ($Q_{\text{ID}}$) interact with the joint audiovisual feature sequences ($F^*$) in each transformer block, where the latter serve as the Key ($K$) and Value ($V$).
This yields two distinct sets of identity tokens: $T^{\tgt}_{\ID} = \texttt{\idmodule}(Q_{\ID}, F^{\tgt})$ from the target features, and $T^{\refer}_{\ID} = \texttt{\idmodule}(Q_{\ID}, F^{\refer})$ from the reference features.
Through this process, the queries adaptively attend to identity-salient regions within the audiovisual sequence while ignoring transient factors.
\vspace{-.5em}

\subsection{Matching cross-modal biometrics}
\vspace{-.5em}
Our ID matching module is designed to evaluate cross-modal biometrics by quantifying the consistency between the target and reference samples.
While the preceding \idmodule extracts individual biometric signatures, this module performs a cross-sample alignment to detect identity mismatches.
We employ $M$ stacked cross-attention blocks where the target tokens (\ie, $T^{\tgt}_{\ID}$) serve as the Query, and the reference tokens (\ie, $T^{\refer}_{\ID}$) act as the Key and Value. 
This matching process generates reference-aware identity tokens, $\bar{T}^{\tgt}_\ID$, which not only retain the intrinsic identity characteristics of the target but also embed information about similarities and discrepancies with the reference. 
As a result, even subtle inconsistencies between the two identities can be amplified and clearly manifested in the final representation.

To ensure the ID matching module prioritizes identity-relevant cues, we design an auxiliary identity verification task to discern whether the target and reference samples originate from the same individual.
Specifically, the reference-aware identity tokens $\bar{T}_{\ID}^{\tgt}$ are transformed into a global identity feature via average pooling, which is then fed into an MLP head to predict the identity matching logit, $\hat{Y}_{\ID}$. 
This process is supervised by an identity matching loss $\mathcal{L}_{\ID}$, implemented as a binary cross-entropy between the prediction and the ground-truth label $Y_{\ID} \in \{0, 1\}$.
Crucially, any forged content, regardless of its intended target, is treated as a distinct identity from the reference speaker.
Therefore, a pair is assigned $Y_{\ID}$=1 exclusively when both samples originate from distinct, authentic recordings of the same individual.
This auxiliary objective encourages the matching module to learn features that are not only conducive to the final deepfake detection task but also explicitly optimized for robust cross-modal identity verification.

\subsection{Audiovisual deepfake detection}
\ours performs deepfake detection by jointly leveraging fine-grained audiovisual correspondence cues and identity consistency information between the target and the reference.
Subsequently, $\bar{T}^{\tgt}_\ID$ are concatenated with the [CLS] token $T_{\text{[CLS]}}$ and the target audiovisual features $F^{\tgt}$: 
\begin{align}
    X^{\tgt} = [T_{\text{[CLS]}}; \bar{T}^{\tgt}_\ID;F^{\tgt}].
\end{align}
The resulting sequence $X^{\tgt}$ is passed through joint audiovisual transformer blocks (shortly, AV-Transformer), which integrate temporal audiovisual factors (\eg, artifacts, synchronization cues) with identity-centric cues. 
Finally, the refined representation of the [CLS] token from the last transformer layer is fed into a deepfake classifier to produce the final deepfake prediction.

The final training objective, $\mathcal{L}_{\text{all}}$, is composed of a deepfake classification loss $\mathcal{L}_{\RF}$ and the identity matching loss $\mathcal{L}_{\ID}$:
\begin{equation}
    \mathcal{L}_{\text{all}} = \mathcal{L}_{\RF} + \mathcal{L}_{\ID},
\end{equation}
where $\mathcal{L}_{\RF}$ denotes the standard cross-entropy loss for distinguishing real from forged content.
By jointly optimizing these tasks, the model not only learns to detect generic manipulation artifacts but also develops a robust identity-discriminative capability, which in turn serves as a powerful prior for high-fidelity deepfake detection.
\section{Experiments}



\definecolor{myorange}{RGB}{255, 208, 0} 
\definecolor{mygreen}{RGB}{46, 75, 191} 

\begin{table}[t]
\centering
\small
\caption{Cross-dataset performance comparison on FF++. $\dagger$ denotes the reimplemented results based on their official code. $\ddagger$ tests on the publicly available trained weights.}
\vspace{-.7em}
\resizebox{\columnwidth}{!}{ %
\begin{tabular}{lccccc}
\toprule
{Method} & {Modality} & {w/ Ref.} & {Train set} & {AUC} & {AP} \\ 
\midrule


Xception~\cite{rossler2019faceforensics++}$\dagger$ 
& {\color{myorange}{V}} & \xmark & FakeAVCeleb 
& 59.48 & 76.60 \\

ID-Reveal~\cite{idreveal}$\ddagger$ 
& {\color{myorange}{V}} & \cmark & VoxCeleb2 
& \textbf{81.28} & 88.19 \\
\midrule
AVAD~\cite{feng2023self}$\ddagger$   
& {\color{mygreen}{AV}} & \xmark & LRS2/LRS3 
& 67.01 & \underline{88.70}\\

POI-Forensics~\cite{poiforensics}$\ddagger$ 
& {\color{mygreen}{AV}} & \cmark & VoxCeleb2 
& 51.33 & 84.75 \\

\midrule
\ours                        
& {\color{mygreen}{AV}} & \cmark & FakeAVCeleb 
& \underline{79.78} & \textbf{91.00} \\
\bottomrule
\end{tabular}
} %
\label{tab:ff++}
\end{table}

\definecolor{myorange}{RGB}{255, 208, 0}
\definecolor{mygreen}{RGB}{46, 75, 191}




\begin{table}[t]
\centering
\small
\caption{Cross-dataset performance comparison on KoDF. $\ddagger$ denotes results evaluated using publicly available pretrained weights.}
\vspace{-.7em}
\resizebox{\columnwidth}{!}{
\begin{tabular}{@{}lccccc@{}}
\toprule
{Method} & {Modality} & {w/ Ref.} & {Train set} & {AUC} & {AP} \\ 
\midrule
Xception~\cite{rossler2019faceforensics++}       
& {\color{myorange}{V}}  & \xmark & FakeAVCeleb & 77.7  & 76.9  \\

LipForensics~\cite{haliassos2021lips}   
& {\color{myorange}{V}}  & \xmark & FakeAVCeleb  & 86.6  & 89.5  \\

FTCN~\cite{zheng2021exploring}           
& {\color{myorange}{V}}  & \xmark & FakeAVCeleb & 68.1  & 66.8  \\

ID-Reveal~\cite{idreveal}$\ddagger$ 
& {\color{myorange}{V}}  & \cmark & VoxCeleb2  & 62.4  & 63.9 \\

RealForensics~\cite{haliassos2022leveraging}  
& {\color{myorange}{V}}  & \xmark & FakeAVCeleb  & 93.6  & 95.7  \\
\midrule

AV-DFD~\cite{zhou2021joint}         
& {\color{mygreen}{AV}} & \xmark & FakeAVCeleb & 82.1  & 79.6  \\

AVAD~\cite{feng2023self}           
& {\color{mygreen}{AV}} & \xmark & LRS2/LRS3 & 86.9  & 87.6  \\

POI-Forensics~\cite{poiforensics}$\ddagger$
& {\color{mygreen}{AV}} & \cmark & VoxCeleb2 & 70.5 & 64.9 \\

AVFF~\cite{oorloff2024avff}              
& {\color{mygreen}{AV}} & \xmark & FakeAVCeleb & 95.5  & 93.1  \\

FRADE~\cite{nie2024frade} 
& {\color{mygreen}{AV}} & \xmark & FakeAVCeleb &  92.4 & - \\

ICSAV~\cite{anshul2025intra} 
& {\color{mygreen}{AV}} & \xmark & LRS2 & \underline{99.2} & \underline{98.6}  \\

FoVB~\cite{nie2025towards} 
& {\color{mygreen}{AV}} & \xmark & FakeAVCeleb &  94.3 & - \\

\midrule

\ours                     
& {\color{mygreen}{AV}} & \cmark & FakeAVCeleb & \textbf{99.4} & \textbf{99.4} \\
\bottomrule
\end{tabular}
}
\label{tab:kodf}
\end{table}

\subsection{Datasets}
\vspace{-.3em}
We train \ours on FakeAVCeleb~\cite{khalid2021fakeavceleb}, which includes both audio and visual manipulations. 
Visual manipulations were created using FaceSwap~\cite{kowalski2018faceswap}, FSGAN~\cite{nirkin2019fsgan}, and Wav2Lip~\cite{prajwal2020lip}, while audio manipulations were generated with SV2TTS~\cite{jia2018transfer}. 
The dataset provides deepfake videos with only visual manipulation, only audio manipulation, or manipulations in both modalities.
We further sourced 37k real videos corresponding to the same 500 identities from VoxCeleb2~\cite{voxceleb2} to use as reference videos.
Following prior works~\cite{yang2023avoid, oorloff2024avff}, FakeAVCeleb is split 70\% for training, and 30\% for a test set.
To assess cross-dataset generalization, we evaluate on FaceForensics++ (FF++)~\cite{rossler2019faceforensics++}, encompassing five manipulation methods~\cite{faceswap2018, kowalski2018faceswap, thies2016face2face, thies2019deferred, li2019faceshifter} ranging from face-swapping to expression manipulation.
Evaluation is conducted on the official test split for videos where audio tracks could be reliably extracted.
For zero-shot cross-lingual evaluation, we utilize KoDF~\cite{kwon2021kodf}, a Korean speech deepfake dataset manipulated with six different  techniques~\cite{kowalski2018faceswap, perov2020deepfacelab, nirkin2019fsgan, siarohin2019first, yi2020audio, prajwal2020lip}.
Following~\cite{feng2023self, oorloff2024avff}, we randomly sampled 100 real and 100 audio-driven manipulated fake videos for evaluation, except manipulation techniques already employed in FakeAVCeleb.
\vspace{-.3em}

\subsection{Implementation details}
Audiovisual encoder is initialized on Synchformer~\cite{iashin2024synchformer} pretrained on LRS3~\cite{afouras2018lrs3}. 
Videos are sampled at 25 fps, while audio is sampled at 16 kHz and converted into 128-channel mel-spectrograms with 25 ms windows and a 10 ms hop.
From each video, we extract a sequence of 8 overlapping segments with a duration of 0.64s (16 frames). 
The model is optimized by Adam with an initial learning rate of $1e$-$5$, following a cosine annealing scheduler with a linear warmup that decays to $1e$-$6$.
To account for class imbalance, weighted sampling is applied during training.

\subsection{Evaluation protocols}
For both training and evaluation, reference videos are sampled from real videos of the same identity as the target.
Importantly, the real videos used for reference sampling are strictly disjoint between training and evaluation to ensure rigorous evaluation and prevent data leakage.
During evaluation, each target video is partitioned into 2.88s windows with a 5\% temporal overlap.
A single 3s segment is randomly sampled from the reference video and consistently paired with all target windows.
The final video-level prediction is determined by averaging the softmax probabilities across all windows.

\begin{table}[t]
\centering
\small

\caption{Intra-dataset comparison of different unimodal (V) and multimodal (AV) methods on FakeAVCeleb.}
\vspace{-.7em}
\resizebox{0.78\columnwidth}{!}{ %
\begin{tabular}{l c c c c}
\toprule
{Method} & {Modality} & {ACC} & {AUC} \\
\midrule
Xception~\cite{rossler2019faceforensics++}       & \textcolor{myorange}{V}  & 67.9 & 70.5  \\
LipForensics~\cite{haliassos2021lips}   & \textcolor{myorange}{V}  & 80.1 & 82.4  \\
FTCN~\cite{zheng2021exploring}           & \textcolor{myorange}{V}  & 64.9 & 84.0  \\
CViT~\cite{wodajo2021deepfake}           & \textcolor{myorange}{V}  & 69.7 & 71.8  \\
RealForensics~\cite{haliassos2022leveraging}  & \textcolor{myorange}{V}  & 89.9 & 94.6  \\
\midrule
VFD~\cite{cheng2023voice}                & \textcolor{mygreen}{AV} & 81.5 & 86.1  \\
AVoiD-DF~\cite{yang2023avoid}           & \textcolor{mygreen}{AV} & 83.7 & 89.2  \\
AVFF~\cite{oorloff2024avff}               & \textcolor{mygreen}{AV} & 98.6 & 99.1  \\

FRADE~\cite{nie2024frade} 
& {\color{mygreen}{AV}} &   98.6 & \textbf{99.8}  \\

CAD~\cite{du2025cad} & \textcolor{mygreen}{AV} & \underline{99.0} & 99.6 \\

FoVB~\cite{nie2025towards} & \textcolor{mygreen}{AV} & 98.5 & \underline{99.7} \\

\midrule
\ours               & \textcolor{mygreen}{AV} & \textbf{99.2} & \underline{99.7} \\
\bottomrule
\end{tabular}
}%

\label{tab:fakeavceleb}
\end{table}

\begin{table}[t]

\centering
\small
\caption{Effectiveness of the identity matching design.}\label{tab:ablation}
\vspace{-.7em}
\resizebox{0.92\columnwidth}{!}{ %
\begin{tabular}{llccc}
\toprule
 {Method}                       & {ACC}            & {AUC}            & AP             \\ 
    \midrule

     \ours                        & \textbf{99.24} & \textbf{99.71} & \textbf{99.99} \\ 

\hspace{1em} w/o reference identity query & 99.03 & 99.33 & \underline{99.98} \\
\hspace{1em} w/o identity matching loss &      \underline{99.09} & \underline{99.69}         & \textbf{99.99}      \\ 
\bottomrule

\end{tabular}
} %
\end{table}

Following standard protocols~\cite{liu2024lips, oorloff2024avff}, we report accuracy (ACC), average precision (AP), and the area under the ROC curve (AUC). 
Given the extreme class imbalance in datasets like FakeAVCeleb (500 real vs. 21,066 fake videos), we prioritize AUC and AP as they provide threshold-independent assessments of model robustness. 

\subsection{Results}\vspace{-.3em}

\tparagraph{Cross-dataset.}
We evaluate the generalization capability in a cross-dataset setting by testing on unseen datasets without any finetuning. 
As reported in \cref{tab:ff++}, our \ours achieves the highest AP and competitive AUC, demonstrating that our model effectively generalized across unseen manipulation types and domain distribution shifts.
Notably, while several reference-based approaches~\cite{poiforensics, idreveal} rely on the massive VoxCeleb2~\cite{voxceleb2} dataset containing over 1 million utterances to train their identity encoders, \ours is trained using only 14k authentic clips.
Despite this significant disparity in data scale, our framework effectively learns to capture robust biometric signatures, demonstrating that million-scale external datasets are not a prerequisite for learning universal identity priors.
This resilience is further evidenced in the cross-lingual evaluation in \cref{tab:kodf}, where our method surpasses previous SoTAs~\cite{nie2025towards, anshul2025intra} by achieving 99.41\% AUC and 99.47\% AP.
Especially, we observe that the performance of existing reference-based models degrades significantly when encountering cross-lingual scenarios.
By anchoring the detection process to cross-modal consistency rather than low-level pixel anomalies, our approach maintains high discriminative power even when encountering zero-shot linguistic contexts and unseen forgery techniques.

\tparagraph{Intra-dataset.}
\cref{tab:fakeavceleb} shows that our method achieves superior improvements over visual-only and audiovisual baselines. 
Compared to RealForensics~\cite{haliassos2022leveraging}, the strongest visual model, our approach achieves gains of 
 9.3\%p in ACC and 5.1\%p in AUC, demonstrating the limitation of relying solely on visual cues.
 Consistent with the superior performance of current audiovisual detectors~\cite{nie2025towards, du2025cad} over unimodal approaches, our \ours achieves the highest accuracy in intra-dataset evaluation.
Explicitly modeling biometric consistency enables our approach to detect subtle discrepancies overlooked by existing fusion-based methods~\cite{nie2025towards, du2025cad}.

\begin{table}[t]
\centering
\small
\caption{Performance varying the number of identity query tokens, and cross-attention layers in AV-Transformer.}\label{tab:architecture}
\vspace{-.7em}
\resizebox{0.9\columnwidth}{!}{ %
\begin{tabular}{cccccc}
\toprule
 {\# Query Tokens} & {CA Layers} & {ACC} & {AUC} & {AP} \\ 
\midrule
 4  & 1 & 99.06 & 98.90 & \underline{99.96} \\
 4  & 2 & \underline{99.15} & \underline{98.96} & \underline{99.96} \\
 6  & 2 & \textbf{99.24} & \textbf{99.71} & \textbf{99.99} \\
\bottomrule
\end{tabular}
} %
\end{table}

\tparagraph{Ablation study.}
We note that all ablation studies were conducted on FakeAVCeleb~\cite{khalid2021fakeavceleb}.
\cref{tab:ablation} examines the effectiveness of the core components in \ours. 
The removal of the reference identity query leads to a noticeable drop, as excluding identity loss results in a slight decline in robustness.
These findings confirm that both our structural design and identity-aware supervision are essential for achieving optimal performance. 
Moreover, \cref{tab:architecture} reports the performance varying the number of query tokens and cross-attention (CA) layers in AV-Transformer while keeping the IDB layers at two.
Increasing the number of CA layers improves performance across all metrics. 
While increasing the number of layers and queries further improves performance, we observe a clear trade-off with computational cost.
This suggests that our chosen configuration strikes a favorable balance between accuracy and efficiency.

\tparagraph{Analysis.}
Although FakeAVCeleb~\cite{khalid2021fakeavceleb} and FaceForensics++~\cite{rossler2019faceforensics++}
are constructed from YouTube videos and already incorporate real-world degradations like noise, compression artifacts, and varying resolutions, we further evaluate robustness under spatial and temporal perturbations to simulate practical scenarios.
As shown in \cref{tab:robustness}, performance remains stable even under 70\% and 90\% spatial cropping, indicating that the model does not rely on specific facial regions.
Notably, \ours demonstrates remarkable stability to limited temporal information, achieving 99.71\% AUC with just a 1-second reference clip.
Overall, these results confirm that \ours is robust to spatial perturbations and variations in reference length, supporting its applicability in realistic deployment scenarios.

\vspace{-.3em}

\begin{table}[t]
\centering
\small
\caption{Robustness analysis under spatial perturbation (random crop) and varying reference clip length on FakeAVCeleb.}\label{tab:robustness}
\vspace{-.7em}
\resizebox{0.75\columnwidth}{!}{ %
\begin{tabular}{lccc}
\toprule
Input noise & ACC & AUC & AP \\
\midrule
 70\% Random Crop & 98.99 & 99.62 & 99.99 \\
 90\% Random Crop & 99.18 & 99.65 & 99.99 \\
\midrule
1 sec. Ref. clip  & 98.92 & 99.71 & 99.99 \\
2 sec. Ref. clip  & 99.12 & 99.71 & 99.99 \\
3 sec. Ref. clip  & 99.24 & 99.71 & 99.99 \\
\bottomrule
\end{tabular}
}
\end{table}

\section{Conclusion}
This study demonstrates that a reference-guided detection framework, \ours, provides a more generic solution for deepfake detection than conventional artifact-based approaches.
\ours effectively identifies forgeries by detecting structural and biometric inconsistencies within the target stream.
Our experiments show that this reference-to-target cross-modal modeling allows the model to remain invariant across different languages and manipulation types, achieving state-of-the-art results without relying on million-scale external datasets.
Conclusively, by shifting the focus from ephemeral artifacts to explicit cross-modal feature correlation, our work establishes a robust and data-efficient defense against increasingly sophisticated digital manipulations.
\section{Generative AI Use Disclosure}
Generative AI tools were used only for improving language clarity and grammar. All scientific contributions and interpretations are entirely the work of the authors.

\bibliographystyle{IEEEtran}
\bibliography{refs}

\end{document}